# Weather Prediction Using CNN-LSTM for Time Series Analysis: A Case Study on Delhi Temperature Data

**Bangyu Li[1], Yang Qian[2]**

[1] Glasgow College, University of Electronic Science and Technology of China, Chengdu, China
[2] Viterbi School of Engineering, University of Southern California, Los Angeles, California, USA

[1] 2445281903@qq.com
[2] yqian442@usc.edu

**Abstract.** As global climate change intensifies, accurate weather forecasting is increasingly crucial for sectors such as agriculture, energy management, and environmental protection. Traditional methods, which rely on physical and statistical models, often struggle with complex, nonlinear, and time-varying data, underscoring the need for more advanced techniques. This study explores a hybrid CNN-LSTM model to enhance temperature forecasting accuracy for the Delhi region, using historical meteorological data from 1996 to 2017. We employed both direct and indirect methods, including comprehensive data preprocessing and exploratory analysis, to construct and train our model. The CNN component effectively extracts spatial features, while the LSTM captures temporal dependencies, leading to improved prediction accuracy. Experimental results indicate that the CNN-LSTM model significantly outperforms traditional forecasting methods in terms of both accuracy and stability, with a mean square error (MSE) of 3.26217 and a root mean square error (RMSE) of 1.80615. The hybrid model demonstrates its potential as a robust tool for temperature prediction, offering valuable insights for meteorological forecasting and related fields. Future research should focus on optimizing model architecture, exploring additional feature extraction techniques, and addressing challenges such as overfitting and computational complexity. This approach not only advances temperature forecasting but also provides a foundation for applying deep learning to other time series forecasting tasks.

**Keywords:** Time series forecasting; convolutional neural networks (CNN); long short-term memory networks (LSTM).

## 1. Introduction

As global climate change intensifies, accurate weather forecasting becomes increasingly crucial in agriculture, energy management, environmental protection, and daily life. Weather prediction accuracy directly impacts socio-economic development and quality of life. Traditional methods, which primarily rely on physical models and statistical approaches, face limitations in handling complex nonlinear and time-varying features, especially under the increased uncertainty brought by climate change [1].

In recent years, deep learning techniques have shown strong potential in weather prediction, particularly in managing complex nonlinear relationships and time series data [2]. For instance, Convolutional Neural Networks (CNN) effectively capture spatial patterns in weather data, while Recurrent Neural Networks (RNN) and Long Short-Term Memory Networks (LSTM) excel in processing meteorological time series, capturing time-dependence in variables [3]. Despite significant progress in areas like rainfall and temperature trend prediction, single models struggle with high-dimensional, multi-scale weather data. Consequently, researchers are exploring hybrid models combining CNN and LSTM to enhance weather prediction performance [4]. This paper focuses on analyzing and predicting historical temperature data in Delhi, India, using a CNN-LSTM hybrid model. We conducted comprehensive data preprocessing and exploratory analysis, followed by constructing and training the model. The experimental results demonstrate the model's strong performance in prediction accuracy and stability, highlighting its potential in weather forecasting. This research provides insights into the effectiveness of the CNN-LSTM model in meteorological time series data and supports the broader application of deep learning in weather prediction.

## 2. Related Work

In time series forecasting, traditional models like Autoregressive Moving Average (ARMA) and Autoregressive Integrated Moving Average (ARIMA) have long been standard due to their effectiveness in handling linear and stationary data. These models, along with Exponential Smoothing, have seen extensive use in fields such as economic forecasting and energy consumption prediction, particularly with stable, cyclical data [1]. However, their linear assumptions limit their performance when dealing with nonlinear, complex, and time-varying data [2, 3].

The advent of Recurrent Neural Networks (RNNs) and Long Short-Term Memory (LSTM) networks addressed some of these limitations by capturing long-term dependencies in sequential data, proving effective in domains like speech recognition and text generation [3, 4]. Despite their strengths, LSTMs may struggle with intricate local patterns and can be prone to overfitting, especially with limited data or noisy feature [5].Conversely, Convolutional Neural Networks (CNNs), originally dominant in computer vision, have shown promise in feature extraction from one-dimensional time series data. Their application to time series classification, such as ECG signal classification, has demonstrated significant accuracy improvements [6].

To address these issues, hybrid models combining CNNs and LSTMs have been developed, leveraging CNN's local pattern recognition with LSTM's temporal modeling capabilities. This CNN-LSTM architecture has shown effectiveness in tasks like precipitation forecasting, offering improved accuracy and robustness across varying conditions [7]. However, these models also introduce increased computational complexity and risk of overfitting, especially with high-dimensional or noisy data [4].

In weather forecasting, particularly in regions with complex climates like Delhi, deep learning models have been increasingly employed for predicting elements such as temperature, rainfall, and wind speed [8]. This study focuses on developing a CNN-LSTM model for temperature forecasting in Delhi, aiming to capture the complex patterns in the region's climate data more effectively than existing methods. The model's performance will be evaluated using metrics like MSE, with discussions on potential challenges such as overfitting and computational demands.

## 3. Prediction Model

*3.1. Dataset*

The dataset used in this study is the Delhi Temperature Dataset, containing temperature records over a specified period, recorded daily or hourly. It includes features like temperature and weather conditions. The training set consists of 7,300 samples, covering around 20 years of historical data, used to train the CNN-LSTM model for temperature prediction.

## 3.2. Data Preprocessing

Data preprocessing is a critical step in model training. When creating time series data, it was standardized to the range of [-1, 1] using the MinMaxScaler, a method widely recognized for its effectiveness in normalizing data [1]. The mathematical formula is as follows.

$$x' = 2 \cdot \frac{x - \min(x)}{\max(x) - \min(x)} - 1 \tag{1}$$

This step accelerated the model's convergence. Initially, the temperature data is normalized to ensure that all features are on the same scale. We apply Min-Max normalization, scaling the data to a range between -1 and 1, which accelerates model training and improves convergence. Next, missing values in the dataset are addressed, either by imputing with the mean of the corresponding column or by removing the affected samples [2]. Additionally, feature engineering is performed to generate time-related features (e.g., season, month) that enhance the quality of the input data for the model [5].

## 3.3. CNN-LSTM Model

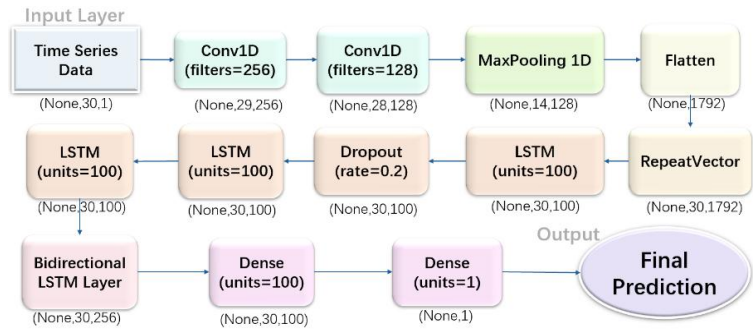

**Figure 1.** Structure of CNN-LSTM Model

The proposed model combines the strengths of one-dimensional convolutional neural networks and long short-term memory networks (LSTM) to predict future temperatures. This architecture in Figure 1 effectively captures both local features [6] and long-term dependencies [3] in time series data, enhancing prediction accuracy. The structure's parameters are presented in Table 1.

**Table 1.** All the layers with specific parameters.

| Layer | Param |
|---|---|
| Conv1D(filters=256) | 768 |
| Conv1D(filters=128) | 65664 |
| LSTM (units=100) | 756800 |
| LSTM (units=100) | 80400 |
| LSTM (units=100) | 80400 |
| LSTM (units=100) | 80400 |
| Bidirectional LSTM | 235520 |
| Dense (units=100) | 25700 |
| Dense (units=1) | 101 |

### A. Convolutional Layer

The first Conv1D layer employs 256 convolutional filters, each with a kernel size of 2. Utilizing the ReLU activation function, this layer introduces non-linearity to the model, allowing it to efficiently extract local features from the input sequence. The input consists of 30-time steps, each containing one feature, making this layer crucial for capturing initial patterns in the data. The second Conv1D layer continues the feature extraction process with 128 convolutional filters. By further refining and extracting higher-level features, the ReLU activation function ensures the model's ability to capture more complex patterns within the time series data.

*B. RepeatVector Layer*

Next, the RepeatVector layer duplicates the flattened feature vector 30 times. This repetition ensures that the features extracted from the previous layers are available at each time step in the subsequent LSTM layers, facilitating the model's ability to learn and process temporal dependencies effectively.

*C. Bidirectional LSTM Layer*

The model features a series of LSTM layers designed to capture complex temporal dependencies within the input sequence. The initial LSTM layer, with 100 units, captures long-term dependencies and returns outputs for each time step, which subsequent layers build upon. To prevent overfitting, a Dropout layer is applied, randomly dropping 20% of neurons during training. Additional stacked LSTM layers further enhance the model's ability to understand intricate patterns in the data. A Bidirectional LSTM layer is also included, processing data in both forward and backward directions to capture dependencies from both perspectives, thereby improving predictive performance

*3.4. Model Training and Optimization*

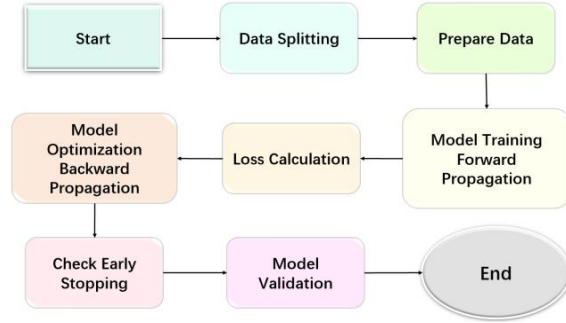

**Figure 2.** Process of model training and optimization

The model training process (see Figure 2) begins with splitting the dataset into training and testing sets to ensure the model's generalization ability. Typically, the data is divided proportionally, allowing the model to be validated on unseen data. In this study, the temperature data from the previous 30 days is used to predict the temperature of the next day. The training steps include forward and backward propagation from equation (1) (2) (3) and (4), where the model parameters are adjusted by minimizing the loss function $\mathcal{L}(\theta)$. EarlyStopping callback is employed to halt training when the training loss does not decrease over seven consecutive epochs, thereby preventing overfitting.

$$m_t = \beta_1 m_{t-1} + (1 - \beta_1)\nabla_\theta \mathcal{L}(\theta_t) \tag{2}$$

$$v_t = \beta_2 v_{t-1} + (1 - \beta_2)\left(\nabla_\theta \mathcal{L}(\theta_t)\right)^2 \tag{3}$$

$$\hat{m}_t = \frac{m_t}{1 - \beta_1^t}, \quad \hat{v}_t = \frac{v_t}{1 - \beta_2^t} \tag{4}$$

$$\theta_{t+1} = \theta_t - \eta \frac{\hat{m}_t}{\sqrt{\hat{v}_t} + \epsilon} \tag{5}$$

Where: $m_t$ and $v_t$ are the first and second moment estimates., $\beta_1$ and $\beta_2$ are the exponential decay rates for these estimates., $\eta$ is the learning rate., $\epsilon$ is a small constant to prevent division by zero.

## 4. Results

*A. Loss Functions*

The choice of loss function has a significant impact on the prediction accuracy of the model. In this experiment, we use both Mean Squared Error (MSE) and Root Mean Squared Error (RMSE) from equation (6) and (7) as loss functions below to evaluate the model's performance. MSE is commonly used to measure the discrepancy between predicted and actual values [7], making it suitable for tasks where we aim to minimize large errors. RMSE, on the other hand, provides a more interpretable metric by taking the square root of MSE, which puts more emphasis on larger errors. The combination of these two loss functions [8] helps in obtaining a more robust evaluation of the model's predictive capabilities. Recent studies [9], [10], and [11] support the effectiveness of these methodologies in various temperature prediction tasks.

$$\text{MSE} = \frac{1}{n} \sum_{i=1}^{n} (y_i - \hat{y}_i)^2 \tag{6}$$

$$RMSE = \sqrt{MSE} \tag{7}$$

*B. Model Training Strategy*

During model training, hyperparameters such as learning rate $\eta$ and batch size $B$ are fine-tuned to optimize the model's performance. The learning rate $\eta_t$ can be dynamically adjusted using learning rate schedules in equation (8).

$$\eta_t = \eta_0 \cdot \frac{1}{1 + \text{decay} \cdot t} \tag{8}$$

The training process begins with an initial learning rate, denoted as $\eta_0$, and a decay factor applied over time, where $t$ represents the current epoch or iteration. The maximum number of training epochs, $T$, is set to 300. However, this number may be reduced through an early stopping mechanism. Early stopping is triggered if the validation loss ($\mathcal{L}_{val}$) fails to improve over a predefined number of epochs, known as the patience period ($p$). Specifically, training is halted if $\mathcal{L}_{val}$ at epoch $t$ is greater than the minimum $\mathcal{L}_{val}$ observed during the previous $p$ epochs. Throughout the training process, detailed logs are maintained to monitor progress.

*C. Model Performance Evaluation*

We plot the change curve of the loss function during the training process to observe the model's convergence. The results indicate that both loss functions decrease rapidly in the early stage of training and then stabilize, suggesting that the model successfully learns the patterns in the data, which improves the prediction accuracy.

To comprehensively evaluate the model's performance, we calculate the mean square error (3.26217) and root mean square error (1.80615). The results demonstrate that the CNN-LSTM model performs well on these metrics, with high prediction accuracy and stability. The prediction curve's high similarity to the test curve further supports the model's effectiveness in temperature prediction.

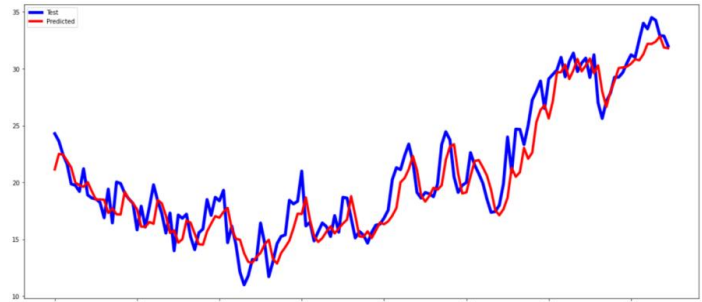
**Figure 3.** The prediction curve and test curve.

## 5. Conclusion

In this study, we employed an optimized CNN-LSTM model to predict temperature time series in Delhi, India. The model effectively captures complex spatio-temporal features, significantly improving prediction accuracy. By thoroughly preprocessing and analyzing historical meteorological data from 1996 to 2017, the CNN-LSTM model leverages the feature extraction capabilities of CNNs and the time series processing strengths of LSTMs. The results show that our model outperforms traditional time series prediction methods, offering higher accuracy and stability.However, the study acknowledges certain limitations, such as the need to improve prediction accuracy under extreme weather conditions and enhance dataset representativeness. Future research should focus on optimizing the model architecture, exploring better feature extraction techniques, and applying the model to a broader range of time series forecasting tasks. Additionally, addressing computational complexities and overfitting through regularization techniques will be crucial for improving model stability and generalization.

In summary, this research offers an effective methodology for temperature prediction in Delhi and lays the groundwork for future investigations in related areas, with the CNN-LSTM framework expected to advance multi-disciplinary time series forecasting applications.